\documentclass[letterpaper, 10 pt, conference]{ieeeconf}
\IEEEoverridecommandlockouts    
\usepackage{graphics}           
\usepackage{times}              
\usepackage{amsmath}            
\usepackage{amssymb}            
\usepackage{graphicx}
\usepackage{algorithm}
\usepackage[noend]{algpseudocode}
\usepackage{booktabs}
\usepackage{color}
\usepackage{subfigure}
\usepackage{multirow}
\usepackage{setspace}
\definecolor{instructioncolor}{rgb}{.5,.5,.5}
\usepackage{threeparttable}

\makeatletter
\renewcommand{\maketag@@@}[1]{\hbox{\m@th\normalsize\normalfont#1}}%
\makeatother

\usepackage[font=small]{caption}

\def\secref#1{Sec.~\ref{#1}}
\def\figref#1{Fig.~\ref{#1}}
\def\tabref#1{Tab.~\ref{#1}}
\def\eqref#1{Eq.~(\ref{#1})}

\makeatletter
\usepackage{xspace}
\DeclareRobustCommand\onedot{\futurelet\@let@token\@onedot}
\def\@onedot{\ifx\@let@token.\else.\null\fi\xspace}

\def\etal{{et al}\onedot}
\makeatother

\def\etalcite#1{\etal~\cite{#1}}

\usepackage{array}
\newcolumntype{L}[1]{>{\raggedright\let\newline\\\arraybackslash\hspace{0pt}}m{#1}}
\newcolumntype{C}[1]{>{\centering\let\newline\\\arraybackslash\hspace{0pt}}m{#1}}
\newcolumntype{R}[1]{>{\raggedleft\let\newline\\\arraybackslash\hspace{0pt}}m{#1}}

\newcommand{\wang}[1]{{\textcolor{black}{#1}}}

\newcommand{\RR}{\mathbb{R}}

\newcommand{\m}[1]{{\mbox{{\sffamily\slshape{#1\/}}}}}

\newcommand{\tr}[0]{\sf T}              

\newcommand{\mM}{\m M}

\newcommand{\mQ}{\m Q}

\newcommand{\mS}{\m S}

\usepackage{hyperref}
\hypersetup{
    colorlinks=true,
    citecolor=blue,
    linkcolor=magenta,
    filecolor=magenta,      
    urlcolor=magenta,
}

\title{\LARGE \bf Leveraging Semantic Graphs for Efficient and Robust LiDAR SLAM}

\author{Neng Wang$^1$ \and Huimin Lu$^1$ \and Zhiqiang Zheng$^1$  \and Yun-Hui Liu$^2$ \and Xieyuanli Chen$^1$
  \thanks{$^1$N. Wang, H. Lu, Z. Zheng, X. Chen are with the College of Intelligence Science and Technology and the National Key Laboratory of Equipment State Sensing and Smart Support, National University of Defense Technology, China. $^2$Y.H. Liu is with the T Stone Robotics Institute and Department of Mechanical and Automation Engineering, the Chinese University of Hong Kong, China.}
  \thanks{Corresponding author: Xieyuanli Chen (xieyuanli.chen@nudt.edu.cn).}
  \thanks{\wang{This work was supported by the National Science Foundation of China under Grant 62403478 and 62203460, Young Elite Scientists Sponsorship Program by CAST (No. 2023QNRC001), as well as Major Project of Natural Science Foundation of Hunan Province under Grant 2021JC0004.}
}%
}

\begin{document}
\maketitle
\thispagestyle{empty}
\pagestyle{empty}

\begin{abstract}
    Accurate and robust simultaneous localization and mapping (SLAM) is crucial for autonomous mobile systems, typically achieved by leveraging the geometric features of the environment. Incorporating semantics provides a richer scene representation that not only enhances localization accuracy in SLAM but also enables advanced cognitive functionalities for downstream navigation and planning tasks. Existing point-wise semantic LiDAR SLAM methods often suffer from poor efficiency and generalization, making them less robust in diverse real-world scenarios.
    In this paper, we propose a semantic graph-enhanced SLAM framework, named \mbox{SG-SLAM}, which effectively leverages the geometric, semantic, and topological characteristics inherent in environmental structures.
    The semantic graph serves as a fundamental component that facilitates critical functionalities of SLAM, including robust relocalization during odometry failures, accurate loop closing, and semantic graph map construction.
    Our method employs a dual-threaded architecture, with one thread dedicated to online odometry and relocalization, while the other handles loop closure, pose graph optimization, and map update.  
    This design enables our method to operate in real time and generate globally consistent semantic graph maps and point cloud maps.
    We extensively evaluate our method across the KITTI, MulRAN, and Apollo datasets, and the results demonstrate its superiority compared to state-of-the-art methods. Our method has been released at \href{https://github.com/nubot-nudt/SG-SLAM}{https://github.com/nubot-nudt/SG-SLAM}.
\end{abstract}

\section{Introduction}
\label{sec:intro}
SLAM is a key capability for autonomous navigation systems, such as mobile robots and autonomous vehicles. 
It has experienced substantial advancements over the past few decades and remains an active area of research.
Leveraging accurate range measurements and robustness to illumination changes, LiDAR-based SLAM has become a cornerstone technology in autonomous systems. Mainstream LiDAR SLAM methods mainly rely on geometric scene features to construct point cloud maps~\cite{zhang2014rss,shan2018iros,wang2021iros,pan2021icra,Yu2024arxiv}. Additionally, researchers have explored alternative SLAM paradigms, including surfel-based~\cite{behley2018rss}, mesh-based~\cite{ruan2023icra}, and emerging implicit localization~\cite{li2024cvpr} and mapping~\cite{Deng2023iccv} techniques.

\begin{figure}[t]
	\centering
	\includegraphics[width=1\linewidth]{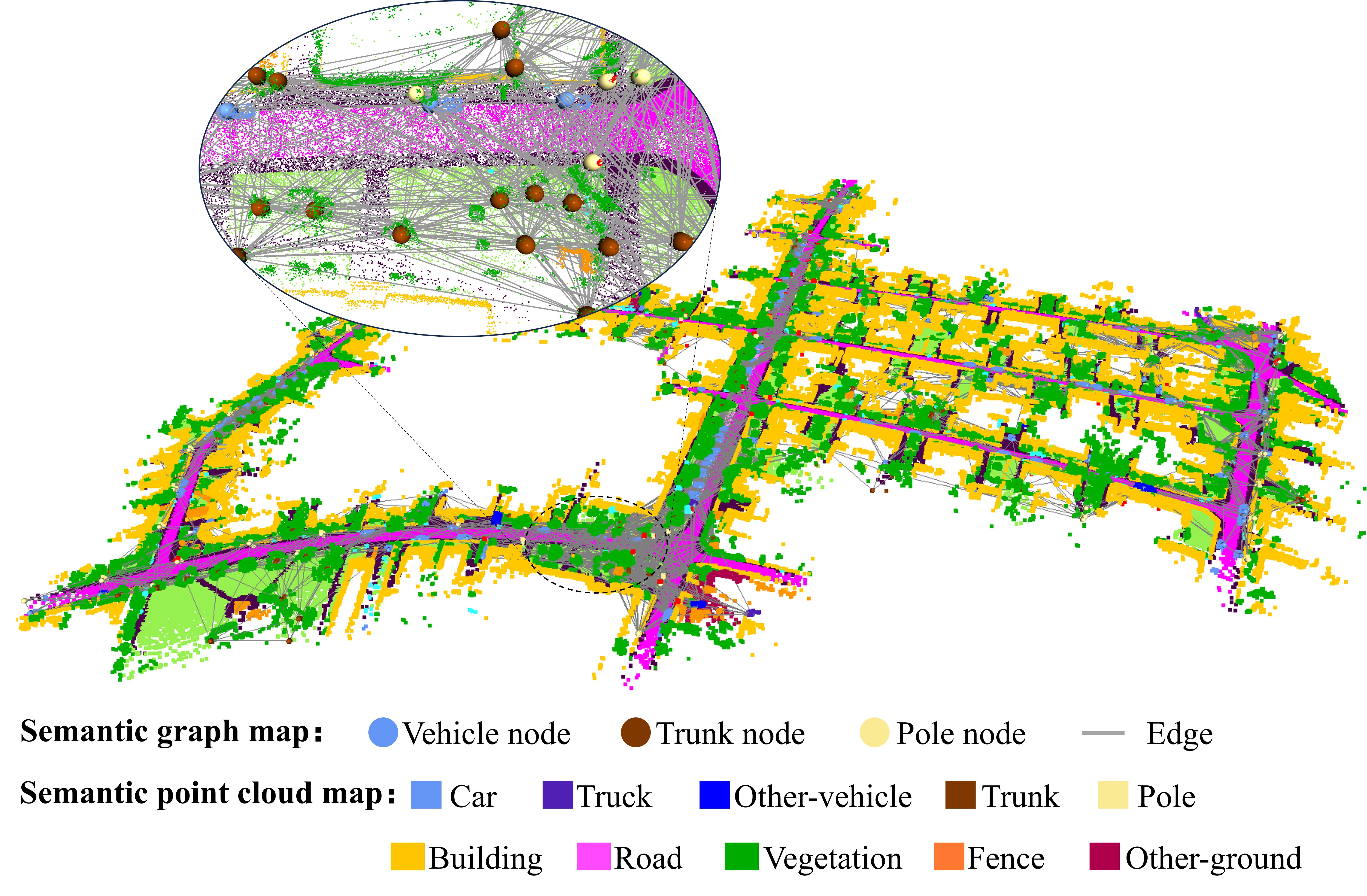}
	\caption{The globally consistent map is constructed through our proposed method, including both the semantic graph map and semantic point cloud map.
	}
	\label{fig:motivation}
\end{figure}

Semantics offer a richer representation of environmental contexts, holding the potential to improve SLAM accuracy and enable more intelligent behaviors in downstream planning and navigation tasks. However, compared to conventional geometry-based approaches, semantic-aware SLAM remains underexplored~\cite{chen2019iros,li2021icra,cui2024icra} due to three key challenges.
First, existing point-wise semantic SLAM methods often lack robustness in adapting to novel environments, particularly when semantic segmentation performance degrades. A widely adopted technique, semantic label consistency checking~\cite{chen2019iros}, verifies label consistency between corresponding points to reject outliers. While this enhances SLAM performance when segmentation is reliable, it becomes problematic when applied to new environments with degraded segmentation accuracy. 
Second, conventional semantic maps, such as semantic point clouds or surfels, are often inefficient for downstream tasks. These representations require substantial storage and cannot be directly utilized for navigation, necessitating additional processing for goal-based path planning.
Third, existing methods lack a recovery mechanism for occasional pose tracking failures, and the lack of open-source robust semantic LiDAR SLAM systems has further hindered progress in the field.

To tackle these limitations, we propose a semantic graph-enhanced SLAM system, named \mbox{SG-SLAM}.
Unlike point-level approaches, \mbox{SG-SLAM} constructs semantic graphs at the object level as the basic elements, inherently improving robustness against point-wise segmentation errors.
In the odometry pipeline, we replace label consistency checking with label-specific weights for residual optimization, mitigating erroneous outlier rejection when segmentation degrades. Additionally, we propose a semantic graph-based relocalization method that enables the system to recover from pose tracking failures, thereby enhancing robustness.
For mapping, our method generates both a globally consistent semantic graph map and a dense point cloud map, as illustrated in~\figref{fig:motivation}. The semantic graph representation significantly improves storage efficiency while maintaining applicability for global localization~\cite{ma2025ral,yin2024icra} and graph-based navigation~\cite{chenc2022iros,Weerakoon2023ral}, and task-level planning~\cite{Ni2024iros,Agia2022corl}.
This representation extends the utility of semantic SLAM beyond traditional applications.
To ensure efficiency, we adopt a parallel dual-thread architecture: the primary thread exclusively handles front-end operations, including odometric estimation and relocalization processes, while the secondary thread manages back-end functionalities encompassing loop closure detection, pose graph optimization, and map update.
This parallelized framework enables real-time operation while maintaining a lightweight and accessible system design.

In summary, the contributions of our work are threefold:
\begin{itemize}
\item[$\bullet$] We propose a novel semantic graph-enhanced SLAM system, which is capable of generating globally consistent semantic maps and supporting real-time online operation.
\item[$\bullet$] We design a semantic graph-based relocalization method in the odometry pipeline, which can automatically detect odometry failure and perform relocalization to resume pose tracking, thereby enhancing the robustness of the odometry.
\item[$\bullet$] Extensive experiments on multiple datasets demonstrate the superiority of our method compared to both state-of-the-art (SOTA) geometry-based and semantic-aided methods. The implementation of our method is made open-source to benefit the community.
\end{itemize}

\section{Related Work}
\label{sec:related}
In this paper, we mainly focus on LiDAR-only SLAM approaches, which can be categorized into geometry-based and semantic-aided methods, depending on whether semantic information is utilized.

\subsection{Geometry-based odometry and SLAM}
Leveraging its precise range measurement capabilities, 3D LiDAR data provides a rich representation of geometric structural information.
Consequently, many methods have been devised to extract geometric features~\cite{zhang2014rss,shan2018iros,wang2021iros,pan2021icra,behley2018rss,Yu2024arxiv}, such as points and planes, which are subsequently utilized to construct residual constraints for LiDAR pose estimation. 
Furthermore, the extraction of such geometric features provides a significant benefit for SLAM, particularly in facilitating the implementation of bundle adjustment~\cite{liu2023tro,liu2023ral,Yu2024arxiv}, thereby improving mapping performance.
As a representative work, LOAM~\cite{zhang2014rss} extracts edge and planar features from raw point clouds based on point smoothness analysis and matches these feature points with local map to construct point-to-line and point-to-plane residual constraints, which form the foundation for precise odometry estimation.
Building upon LOAM, LeGO-LOAM~\cite{shan2018iros} employs a point cloud segmentation algorithm in advance to filter out noise points, ensuring more accurate feature extraction. Additionally, it incorporates a two-step Levenberg-Marquardt (LM) nonlinear optimization technique for pose estimation, improving computational efficiency.
Subsequently, FLOAM~\cite{wang2021iros} introduces a non-iterative two-step distortion compensation method to replace the inefficient iterative distortion compensation method employed  in~\cite{zhang2014rss,shan2018iros}, further enhancing the algorithm's efficiency and enabling it to operate in real-time.
To improve the robustness of SLAM, Pan~\etalcite{pan2021icra} propose MULLS, which incorporates a broader variety of feature points, enabling it to adapt to more challenging scenarios.
Different from the aforementioned method, Behley~\etalcite{behley2018rss} introduce the surfel feature and develop a comprehensive SLAM system with loop closure detection and map update.

Although feature-based methods exhibit precise performance, the process of feature extraction is not only time-consuming but also necessitates the adjustment of many parameters to adapt to varying environments.
Therefore, some works~\cite{Dellenbach2022icra,vizzo2023ral} focus on dense point Iterative Closest Point (ICP)-based methodologies, attempting to perform point-to-point ICP on down-sampled point clouds directly. This approach obviates the necessity for predefined feature extraction, demonstrating greater robustness.

\subsection{Semantic-aided odometry and SLAM}
Semantic information enables the prior differentiation of various objects within a scene, offering significant potential to improve the efficacy and accuracy of scene recognition and localization processes.
Consequently, some LiDAR SLAM works integrate semantic information to enhance performance.
In its early development, SuMa++~\cite{chen2019iros} employs a dynamic object detection and removal module leveraging semantic information, and introduces semantic label consistency checks into the ICP process. These enhancements collectively improve the SLAM performance compared to the original SuMa~\cite{behley2018rss}.
Similarly, Chen~\etalcite{chen2021icrapsf} integrate semantic information to distinguish between dynamic and static objects and design various parameterized semantic features for odometry estimation.
Li~\etalcite{li2021icra} propose SA-LOAM, a semantic-aided SLAM approach that incorporates semantic information into the LOAM framework and integrates loop closing, enabling more precise pose estimation.
To further improve SLAM accuracy, Jiang~\etalcite{Jiang2023iros} develop a semantic-based LiDAR-Visual SLAM framework. Semantic information is primarily utilized for loop closure detection and the determination of degenerate scenarios.
Recently, Cui~\etalcite{cui2024icra} design semantic-aided odometry, SAGE-ICP, which primarily constructs a semantically adaptive voxel map capable of preserving some scarce but critical semantic information, further improving the localization performance.

The semantic graph, as the cornerstone of our approach, integrates both semantic information and the topological structure of the environment, thereby providing a comprehensive representation of the scene.
It is more frequently employed for place recognition~\cite{kong2020iros,Pramatarov2022iros} and less commonly used for the construction of SLAM systems.
Although SA-LOAM~\cite{li2021icra} incorporates semantic graph representations within its SLAM framework, the implementation of its deep learning-based loop closure detection mechanism presents significant challenges for real-time online operation.
Moreover, its system primarily emphasizes localization accuracy improvement while inadequately addressing critical aspects of map consistency maintenance. 
Specifically, it neglects to update instance attributes through multi-observation fusion and avoids their duplicated representation in the map.
In contrast, our approach not only integrates semantic graphs for robust odometry and loop closure but also ensures map consistency.


\begin{figure*}[ht]
	\centering
	\includegraphics[width=1\linewidth]{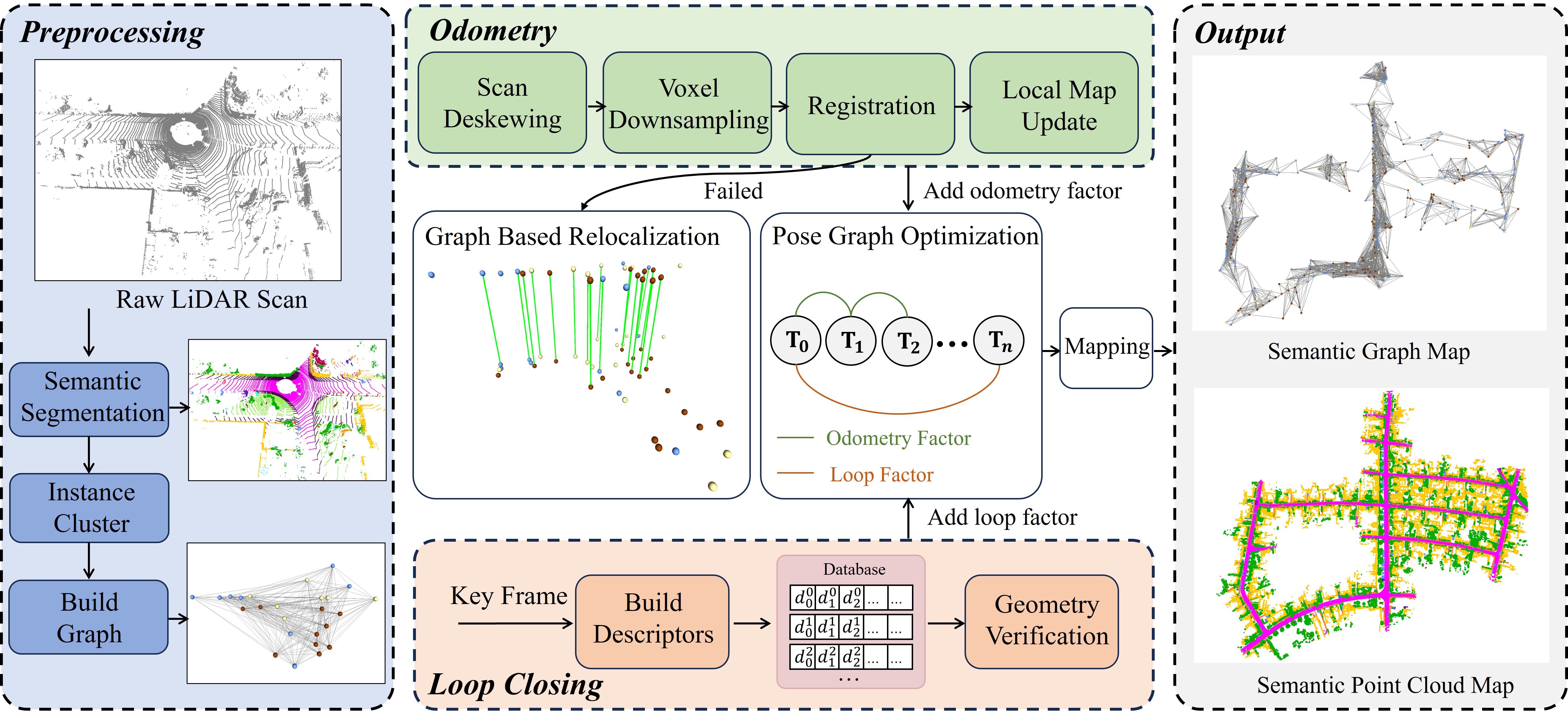}
	\caption{The framework of our SLAM system. All components are partitioned into two distinct threads: one primarily dedicated to data preprocessing and odometry estimation, while the other thread handles back-end operations, including loop closing, pose graph optimization, and map maintenance.
	}
	\label{fig:pipeline}
	\vspace{-0.5cm}
\end{figure*}

\section{Semantic Graph SLAM}
\label{sec:slam}
The proposed framework, as illustrated in~\figref{fig:pipeline}, comprises several key components: LiDAR data preprocessing (~\secref{sec:graph}), odometry estimation and relocalization (~\secref{sec:front_end}), loop closing (~\secref{sec:loop}), pose graph optimization and map maintenance (~\secref{sec:mapping}). 
We will detail each component in subsequent subsections. 

\subsection{Preliminaries}
\label{sec:graph}
Given a raw LiDAR scan $\mS$, we assume semantic labels are obtainable from existing segmentation methods. Notably, we deliberately avoid integrating specific segmentation approaches into our SLAM system to maintain its lightweight nature and ensure extensibility across different semantic segmentation frontends. 
We can build the semantic graph $\mathcal{G} = \{ V, E\}$ using our prior work~\cite{wang2024ral}, where $V$, $E$ represent the instance node set and the edge set, respectively.
Each instance node in the semantic graph is associated with a descriptor that facilitates instance matching and tracking.

\subsection{Odometry with Robust Relocalization}
\label{sec:front_end}

\subsubsection{Odometry Estimation}
\label{sec:odometry}
Odometry aims to estimate the trajectory of a moving LiDAR sensor. 
Following~\cite{Dellenbach2022icra,vizzo2023ral}, we employ the ICP-based method to sequentially register the LiDAR scan.
Prior to registration, we first perform scan deskewing~\cite{vizzo2023ral} to compensate for measurement inaccuracies caused by sensor motion. 
Concurrently, for computational efficiency considerations, we implement voxel-based downsampling, with a voxel size of $v$, thereby yielding a downsampled point cloud 
$\mS_t=\{\mathbf{p}_{i}|\mathbf{p}_i \in \RR^4\}, \mathbf{p}_i=[x_i,y_i,z_i,l_i]^{\tr}$ with semantic label.
We transform $\mS_t$ into the global coordinate frame using the widely adopted constant-velocity motion prediction model $\mathbf{T}_{\text{p,t}}$
\begin{equation}
\mS'_t = \{\mathbf{p}'_{i} = \mathbf{T}_{\text{t-1}} \mathbf{T}_{\text{p,t}} \mathbf{p}_i \},
\end{equation} 
\begin{equation}
\mathbf{T}_{\text{p,t}} = \mathbf{T}^{-1}_{\text{t-2}} \mathbf{T}_{\text{t-1}},
\end{equation} 
where $\mathbf{T}_{\text{t-2}}$,$\mathbf{T}_{\text{t-1}}$ demotes the LiDAR pose in the past time steps, respectively.
Note that we replace $l_i$ with 1 in actual calculations.
Subsequently, we register  $\mS'_t$ to local point cloud map $\mQ_p = \{\mathbf{q}_i| \mathbf{q}_i \in \RR^4\}$ to derive the odometry estimates.
Furthermore, we argue that different semantic objects contribute differently to the localization process. Pole-like objects inherently offer greater assistance in improving localization accuracy. Therefore, we assign different weights to the residual optimization.
\begin{equation}
\label{eq:registration}
\mathbf{T}_{\text{t}} = \mathop{\text{min}}\limits_{\mathbf{T}} \sum\limits_{(\mathbf{p}',\mathbf{q})\in \mathcal{C} } \text{w}(l)\Vert \mathbf{T} \mathbf{p}' - \mathbf{q}  \Vert_{2},
\end{equation} 
where $\mathcal{C}$ is point-to-point correspondence between $\mS'_t$ and $\mQ_p$ established by searching for the nearest neighbors. Like~\cite{vizzo2023ral}, we employ a hash-based voxel map to accelerate the search process.
We iteratively apply~\eqref{eq:registration} to achieve optimal registration, terminating the process once the correction error falls below the threshold $\gamma$. Here, $\text{w}$ represents a predefined weight assigned to different labels. Unlike the rigid constraints imposed by label consistency checks~\cite{chen2019iros}, which risk indiscriminately rejecting a significant number of inliers when semantic segmentation performance deteriorates, our weighted residual optimization incorporates a soft constraint, thereby improving the robustness of the registration process.

\subsubsection{Local Map}
\label{sec:local_map}
We construct two distinct local map representations: a dense semantic point cloud map and a topological semantic graph map. The former is used for frame-to-map registration to derive odometry estimates, while the latter enables efficient relocalization capabilities and supports the generation of a globally consistent graph-based map representation.
The local point cloud map $\mQ_p$ encompasses point-wise coordinates as well as its associated semantic label.
Upon obtaining the odometry estimation, we transform $\mS_t$ into a global coordinate and add it into the local map.
For memory-efficient and fast nearest neighbor search in the registration, we utilize voxel to store a subset of points and construct a hash table to store these voxels~\cite{vizzo2023ral}.
Each voxel can store a maximum of $N_\text{max}$ points, and voxels whose position exceeds a distance of $d_\text{max}$ from the current LiDAR position are deallocated.

Local semantic graph map $\mQ_g = \{V',E'\}$ comprises all instance nodes within the local mapping region and their corresponding connecting edges. 
This representation fundamentally differs from the local point cloud map maintenance in its requirement for continuous node tracking, which is essential for maintaining instance node consistency across the map.
As mentioned in~\secref{sec:graph}, each instance node encompasses a descriptor, which is initially generated during the construction of the semantic graph.
In order to quickly find the node correspondence between the current semantic graph $\mathcal{G}_t$ and  $\mQ_g$, we employ a KD-tree constructed from all node descriptors within the local graph map, subsequently utilizing each instance node descriptor from the $\mathcal{G}_t$ as query vectors to retrieve the similar candidates.
Furthermore, to ensure the accuracy of the candidates, we employ an outlier pruning algorithm proposed in our prior research~\cite{wang2024ral} to reject incorrect node correspondences.
Once the tracking of instance nodes is achieved, we proceed to update their attributes within the local graph map, including 3D coordinates and bounding box.
Instance nodes in the  $\mathcal{G}_t$ that lack a corresponding match in $\mQ_g$ will be regarded as new nodes and subsequently inserted into the local graph map.
Following this, the descriptor for each node in the $\mQ_g$ is re-computed based on their adjacency relationships~\cite{wang2024ral} to facilitate the next round of node matching.

\subsubsection{Relocalization}
\label{sec:relocation}
For semantic SLAM, robustness is critically essential. Degradation in the performance of the front-end semantic segmentation module or occasional sensor malfunctions can significantly increase the risk of failure in odometry registration.
Existing methods~\cite{chen2019iros,cui2024icra} often fail to address such scenarios, resulting in the breakdown of the odometry system.
To tackle this issue, we devise a relocalization mechanism capable of automatically detecting odometry errors and executing relocalization, thereby enabling the robot to resume pose tracking.

Specifically, if the odometry is functioning correctly, the discrepancy between the initial estimate provided by the constant velocity motion model and the final frame-to-map registration poses should remain below a certain threshold. Conversely, a deviation beyond this threshold is indicative of potential inaccuracies within the odometric estimations.
\begin{equation}
\mathbf{T}_{\text{error}} =( \mathbf{T}_{\text{t-1}} \mathbf{T}_{\text{p,t}})^{-1}\mathbf{T}_{\text{t}}.
\end{equation} 

$\mathbf{T}_{\text{error}} = (\mathbf{R}_{\text{error}} ,\mathbf{t}_{\text{error}} )$, represented by a rotation matrix $\mathbf{R} \in SO(3)$ and a translation vector $\mathbf{t} \in \RR^3$. If the distance error $\text{norm}(\mathbf{t}_{\text{error}})>t_o$ or the angular error exceeds $r_o$, the odometry is deemed inaccurate and subsequent relocalization is executed, where $\text{norm}(\cdot)$ denote the L2 norm function.

In the~\secref{sec:local_map}, we have already obtained the instance node correspondences between the $\mathcal{G}_t$ and local graph map $\mQ_g$, denoted $\mM=\{(\mathbf{v}^c_1,\mathbf{v}^g_1), (\mathbf{v}^c_2,\mathbf{v}^g_2),..., (\mathbf{v}^c_m,\mathbf{v}^g_m) \}$. $\text{v}^c$ represents the instance coordinates in the $\mathcal{G}_t$, while $\text{v}^g$ denotes the coordinates of the matched instance in the $\mQ_g$.
Following this, we directly utilize the RANdom SAmple Consensus (RANSAC) algorithm~\cite{fischler1981acm} and Singular Value Decomposition (SVD)~\cite{besl1992pami} to compute the pose $\mathbf{T}_{\text{r}}$ for relocalization.
We transform the instance node of the $\mathcal{G}_t$ into the local graph map and calculate the inlier ratio:
\begin{equation}
I = \frac{\sum\limits_{(\mathbf{v}^c_i,\mathbf{v}^g_i)\in \mM} \mathbb{I}( \Vert \mathbf{T}_{\text{r}} \cdot \mathbf{v}^c_i -\mathbf{v}^g_i \Vert_2 < \tau )}{m},
\end{equation} 
where $\tau$ denotes the inlier distance threshold and $\mathbb{I}(\cdot)$ is an indicator function for which the statement is true. 
The inlier ratio $I$ exceeding the predefined threshold $I_r$ indicates a successful relocalization. 
This relocalization method does not rely on an initial pose, making it highly efficient and robust.
Upon successful relocalization, $\mathbf{T}_\text{r}$ is employed as the initial value to perform dense point-based ICP, ensuring precise pose estimation and resuming pose tracking.

\subsection{Loop Closing}
\label{sec:loop}
Loop closing can effectively eliminate the accumulated errors in odometry, making it an indispensable component of SLAM.
We integrate our prior semantic graph-based loop closing method~\cite{wang2024ral} into this SLAM system.
Specifically, each past LiDAR scan is encoded into a descriptor vector, thereby constructing a comprehensive database, while the descriptor of the current scan serves as a query to retrieve potential loop closure candidates.
Following this, geometric verification is performed to determine whether a true loop closure has occurred.
If the similarity of the semantic graph and background points between the query frame and the candidate frame exceeds threshold $l_g$ and $l_b$ respectively, we further estimate their loop closure transformation and add a loop factor to the pose graph for subsequent pose optimization.
We employ a KD-tree to manage the database, and geometric verification is only conducted when the Euclidean distance between the query descriptor and the retrieved descriptor is less than $l_v$.
In real applications, using all LiDAR scans for loop closing is computationally expensive and redundant.
Therefore, we strategically select keyframes at intervals of every $n$ LiDAR scan and only utilize these keyframes for loop closure detection.
This can also be readily replaced with other keyframe selection strategies, such as those based on rotational or translational movement.

\subsection{Mapping}
\label{sec:mapping}
To construct a globally consistent map, we employ a pose graph optimization framework, i.e., gtsam~\cite{Dellaert2012Factor}, to build a factor graph model for managing the LiDAR poses.
It primarily comprises two types of factors: odometry factors and loop closure factors.
After each LiDAR scan is registered, we add an odometry factor and the associated edge into the factor graph, where the edge denotes a relative pose constraint between the continuous odometry factor.
Note that the successfully relocalized LiDAR scan is still regarded as an odometry factor inserted into the factor graph.
Upon detection of a loop closure, the corresponding loop constraint is inserted into the factor graph.
Subsequently, we utilize the Incremental Smoothing and Mapping (iSAM2)~\cite{Kaess2012ijrr} to optimize the factor graph, thereby refining the poses of all LiDAR scans.

Our global map comprises a global point cloud map and a semantic graph map.
In consideration of memory efficiency, the global point cloud map is no longer updated in real-time. Instead, it is only generated after all the LiDAR scan poses have been fully computed.
The semantic graph map preserves the attributes of all instance nodes, encompassing their coordinates, labels, and associated edges.
Similar to the local semantic graph map update, each newly added instance node, that is, one that has not been successfully tracked in the local map, will be inserted into the global graph map. Concurrently, the attributes of successfully tracked instance nodes are correspondingly updated.
When a loop closure occurs, instance nodes of the current scan that successfully match within the loop scan will not be reinserted as new instance nodes into the global map.
This practice maintains the consistency of instance nodes within the global map.

\section{Experimental Evaluation}
\label{sec:exp}

%
%

\subsection{Experimental Setup}
\textbf{Datasets.} 
We evaluate our method on widely-used outdoor LiDAR datasets: KITTI~\cite{geiger2012cvpr}, MulRAN~\cite{kim2020icra} and Apollo ~\cite{lu2019cvpr}. 
These datasets are collected on different platforms, with varied sensor setups, and in different environments.
The KITTI odometry dataset is recorded using a Velodyne HDL64 LiDAR across a variety of scene types, including urban, country, and highway.
The MulRAN dataset is collected using an Ouster OS1-64 LiDAR in urban areas surrounding the KAIST campus (KA), Daejeon Convention Center (DC), and Riverside (RS).
Due to occlusions from other sensors, the MulRAN LiDAR data loses nearly 70$^{\circ}$ of its FoV, posing a greater challenge for robustness test.
The Apollo dataset is acquired using a Velodyne HDL64 LiDAR mounted vehicle driving through different areas in southern San Francisco Bay, covering various scenes, such as residential areas, urban downtown areas, and highways.
The original Apollo dataset contains a vast amount of data. We utilize a subset proposed in~\cite{chen2022ral}, which comprises 5 sequences with 6,600 scans.

\textbf{Implementation Details.} 
The detailed parameters used in our experiments are listed in~\tabref{tab:parameter}.
For the residual weights $\text{w}(\cdot)$ in the~\eqref{eq:registration}, we assign higher weights to pole-like objects compared to other objects to signify their importance for localization, i.e., $\text{w}(\text{pole-like}) = 1.2$, $\text{w}(\text{else}) = 1$.
For obtaining semantic labels, we utilize a pre-trained SegNet4D~\cite{Wang2024arxiv} model on the SemanticKITTI~\cite{behley2019iccv} dataset to generate point-wise predictions and we mainly use the static vehicle, pole and trunk to build semantic graphs.
The specific details of the semantic graph construction can be found in our previous research~\cite{wang2024ral}.
We maintain consistent parameter settings for both the KITTI and Apollo datasets. For the MulRAN dataset, we primarily modify the loop closure parameter $l_v=0.18$ and $l_b=0.3$.
Additionally, we exclude the vehicle category from the semantic graph construction due to degraded semantic segmentation performance.
It is worth noting that for semantic prediction in the KITTI dataset, we employ cross-validation to ensure that each test sequence is unseen during the training phase. This ensures greater alignment with practical applications.

\begin{table}[t]
	\caption{The parameters list of our approach.}
    \scriptsize 
    \setlength\tabcolsep{2.6pt}
	\centering
	\begin{threeparttable}
	\begin{tabular}{c|lll}
		\toprule
		parameter & value & description   \\
        \midrule
        $v$ & 0.5 &size of voxel downsampling \\
        $N_\text{max}$& 20 &maximum number of points within each voxel in the map\\
        $d_\text{max}$& 100 &range of local map\\
        $\gamma$ & 0.0001& ICP convergence threshold\\
        $t_o$ & 0.12 &distance threshold for performing relocalization \\
        $r_o$ & 0.01 &angle threshold for performing relocalization \\
        $\tau$ & 0.2 &inlier distance threshold for determine relocalization status\\
        $I_r$ & 0.43 &inlier ratio threshold for determine relocalization status\\
        $l_v$ & 0.1 & the descriptors distance threshold for geometry verification \\
        $l_g$ & 0.5 & the graph similarity threshold for loop closing \\
        $l_b$ & 0.58 & the background similarity threshold for loop closing  \\
        $n$ & 5 & the interval to keyframe selection \\
		\bottomrule
	\end{tabular}
	\end{threeparttable}
	\label{tab:parameter}
\end{table}

\begin{table*}[t]
	\caption{SLAM performance comparison on the KITTI dataset with absolute trajectory error (ATE, RMSE [m]).}
	\centering
    \footnotesize
    \setlength\tabcolsep{7.5pt}
	\begin{threeparttable}
	\begin{tabular}{cc|ccccccccccc|cc}
		\toprule
		&Method& 00$^\ast$ & 01 & 02$^\ast$ & 03 & 04 & 05$^\ast$ & 06$^\ast$ & 07$^\ast$ & 08$^\ast$ & 09$^\ast$ & 10 & Mean$^\ast$ & Mean \\
		\midrule
        \multirow{4}{*}{\rotatebox{90}{Geo.-based}} &MULLS& 1.09 & \underline{1.96} & 5.42 & 0.74 & 0.89 & 0.97 &  0.31 & 0.44 & 2.93 & 2.12 & 1.13 & 1.90 & 1.64 \\
        &CT-ICP& 1.68 & 2.25 & 4.06 & 0.67 & 0.67 & 0.76 & 0.34 & 0.40 & 2.52 & \textbf{0.91} & 0.83 & 1.52 & 1.37 \\
        &BALM2& 0.84 & \textbf{1.83} & 5.06 & 0.57 & 0.64 & 0.62 & \textbf{0.21} & 0.30 & 2.59 &  1.48 & \textbf{0.78} & 1.59 & 1.36 \\
        &SLIM& 0.95 & 3.72 & 1.99 & 0.79 & 0.28 & 0.61 & \underline{0.24} & 0.33 & 2.31 & 3.12 & 1.07 & 1.36& 1.40\\
        \midrule
        \multirow{6}{*}{\rotatebox{90}{Sem.-aided}}&SuMa++& 1.19 & 14.25 & 8.99 & 0.99 & 0.31 & 0.63 & 0.49 & 0.37 & 2.68 & 1.27 & 1.32 & 2.23 & 2.95\\
        &SA-LOAM& 0.99 & - & 9.24 & - & - & 0.75 & 0.64 & 0.36 & 3.24 & 1.20 & - & 2.34 & - \\
        &SELVO& 1.06 & - & 4.03 & - & - & 0.44 & 0.63 & 0.38 & 2.75 & \underline{1.13} & - & 1.49 & - \\
        &SAGE-ICP-SGLC$^\dag$& 0.82 & 3.58 & 2.27 &  \underline{0.48} & \textbf{0.26} &  \underline{0.28} & 0.28 & 0.29 &  \underline{2.00} & 1.84 & 0.82 & \underline{1.11}& 1.17\\
        &Ours(RangeNet)& \underline{0.80} &  3.70 & \textbf{1.78} & \textbf{0.43} &  \underline{0.27} & \textbf{0.27} &  0.27 &  \underline{0.28} &  \underline{2.00} & 1.55 & 0.82 & \textbf{0.99} & \underline{1.11}\\
        &Ours(SegNet4D)& \textbf{0.79} & 3.72 &  \underline{1.83} & \textbf{0.43} &  \underline{0.27} & \textbf{0.27} & 0.26 & \textbf{0.27} & \textbf{1.98} & 1.50 & \underline{0.81} & \textbf{0.99} & \textbf{1.10}\\
		\bottomrule
	\end{tabular}
	\begin{tablenotes}
	\footnotesize
	\item $^\ast$ indicates the sequence is with loops and average results on these sequences. $^\dag$ indicates that this method integrates a loop-closure detection approach within an offline PGO framework. - indicates the result is not reported and the method is not open-sourced. We highlight the best results in bold and the second best is underlined.
	\end{tablenotes}

	\end{threeparttable}
	\label{tab:ate_kitti}
\end{table*}

\begin{table*}[t]
	\caption{SLAM performance comparison on the MulRAN dataset (ATE, RMSE [m]). All sequences are with loops.}
	\centering
    \footnotesize
    \setlength\tabcolsep{11pt}
	\begin{tabular}{c|ccccccccc|c}
		\toprule
		Method& KA01 & KA02 & KA03 & DC01 & DC02 & DC03 & RS01 & RS02 & RS03 & Mean \\
		\midrule
        MULLS & 19.72 & 15.26 & 6.93 & 14.95 & 11.31 & 6.00 & 41.25 & 45.05 & 44.18 & 22.74 \\
        HBA & \underline{3.36} & \underline{3.75} & 3.53 & \textbf{5.19} & \underline{3.20} & \underline{2.54} & \underline{8.92} & \textbf{7.94} & \underline{10.26} & \underline{5.41} \\
        \midrule
        SuMa++ & 48.20 & 29.14 & 40.41 & 29.22 & 22.11 & 25.37 & Failed & Failed & Failed & 32.41 \\
        SAGE-ICP-SGLC$^\dag$& 4.93 & 4.70 & \underline{3.42} & 5.92 & 3.93 & 2.97 & 14.23 & 19.28 & 18.24 &  8.62\\
        Ours & \textbf{2.68} & \textbf{2.77} & \textbf{2.64} & \underline{5.48} & \textbf{2.85} & \textbf{2.19} & \textbf{6.28} & \underline{8.52} & \textbf{7.90} & \textbf{4.59}\\
		\bottomrule
	\end{tabular}
	\label{tab:ate_mulran}
\end{table*}

\begin{table}[t]
	\caption{SLAM performance comparison on the Apollo dataset (ATE, RMSE [m]).}
	\centering
    \footnotesize
    \setlength\tabcolsep{6pt}
	\begin{threeparttable}
	\begin{tabular}{c|ccccccccc|c}
		\toprule
		Method & 00$^\ast$ & 01$^\ast$ & 02 & 03 & 04$^\ast$ & Mean \\
        \midrule
        SuMa++& \textbf{0.33} & \textbf{0.13} & 3.37 & 0.35 & 0.36 & 0.91 \\
        SAGE-ICP-SGLC$^\dag$ & 0.41 & 0.19 & 1.65 & 0.24 & 0.26 & 0.55 \\
        Ours & 0.35 & 0.16 & \textbf{1.22} & \textbf{0.19} & \textbf{0.22} & \textbf{0.43} \\

		\bottomrule
	\end{tabular}

	\end{threeparttable}
	\label{tab:ate_apollo}
\end{table}



\begin{table}[t]
	\caption{Odometry performance comparison on the KITTI dataset with relative translational error in $\%$.}
	\centering
    \footnotesize
    \setlength\tabcolsep{15pt}
	\begin{tabular}{c|c|c|c}
		\toprule
		Method& KITTI & MulRAN & Apollo\\
		\midrule
        KISS-ICP& \underline{0.50} & \underline{2.50} & \underline{0.60}\\
        SuMa++  & 0.70 & 6.81 & 0.79\\
        SA-LOAM-LO& 0.76 & - & -\\
        SELVO-LO & 0.75 & - & - \\
        SAGE-ICP& \textbf{0.49} & 3.69 & 0.69\\
        Ours-LO & \underline{0.50} & \textbf{2.49}& \textbf{0.58}\\
		\bottomrule
	\end{tabular}
	\label{tab:odometry}
\end{table}

\subsection{Localization Performance Evaluation}
\subsubsection{SLAM performance evaluation}
We evaluate our method using the Absolute Trajectory Error (ATE) metric and compare it with SOTA SLAM methods, including a) geometry-based method: MULLS~\cite{pan2021icra}, CT-ICP~\cite{Dellenbach2022icra}, BALM2~\cite{liu2023tro}, SLIM~\cite{Yu2024arxiv} and HBA~\cite{liu2023ral}; b) semantic-aided method: SuMa++~\cite{chen2019iros}, SA-LOAM~\cite{li2021icra}, SELVO~\cite{Jiang2023iros}. Additionally, considering the scarcity of existing LiDAR-based semantic SLAM baselines, we utilize an offline pose graph optimization framework to integrate the latest LiDAR-based semantic odometry and loop closure, denoted as SAGE-ICP-SGLC~\cite{cui2024icra, wang2024ral}. 
All the aforementioned methods incorporate loop closure detection.

First, we compare the ATE results on the KITTI odometry dataset.
As shown in~\tabref{tab:ate_kitti}, our method achieves an average ATE of 0.99~m on sequences with loop closures and an average ATE of 1.10~m on all sequences, outperforming all the baseline methods.
Although SAGE-ICP-SGLC also achieves satisfactory results, its functionality is confined to offline execution of odometry estimation, loop closure detection, and pose graph optimization, and it lacks the capability to generate the semantic map.
In contrast, our method not only operates online but also constructs the globally consistent semantic map, thereby highlighting its applicability for online applications in autonomous driving and robotics.
For fairer comparison with some earlier methods~\cite{chen2019iros,li2021icra,Jiang2023iros}, we also report the ATE performance using the semantic labels predicted by RangeNet++~\cite{milioto2019iros}.
Note that SAGE-ICP-SGLC utilizes semantic labels from SegNet4D, which is consistent with our method.
The results indicate that our method shows negligible performance variation, which partially attests to its robustness to the semantic frontend.

We further evaluate our method on the MulRAN dataset. Compared with the KITTI dataset, the MulRAN dataset encompasses more challenging scenarios and longer sequence distances.
The results are shown in~\tabref{tab:ate_mulran}, where our method consistently achieves SOTA performance and significantly outperforms other methods.
Additionally, we also report the results on the Apollo dataset and compare them with two semantic-aided methods: SuMa++ and SAGE-ICP-SGLC. The results in~\tabref{tab:ate_apollo} demonstrate that our method remains the best for semantic SLAM.

\begin{figure*}[t]
	\centering
	\includegraphics[width=1\linewidth]{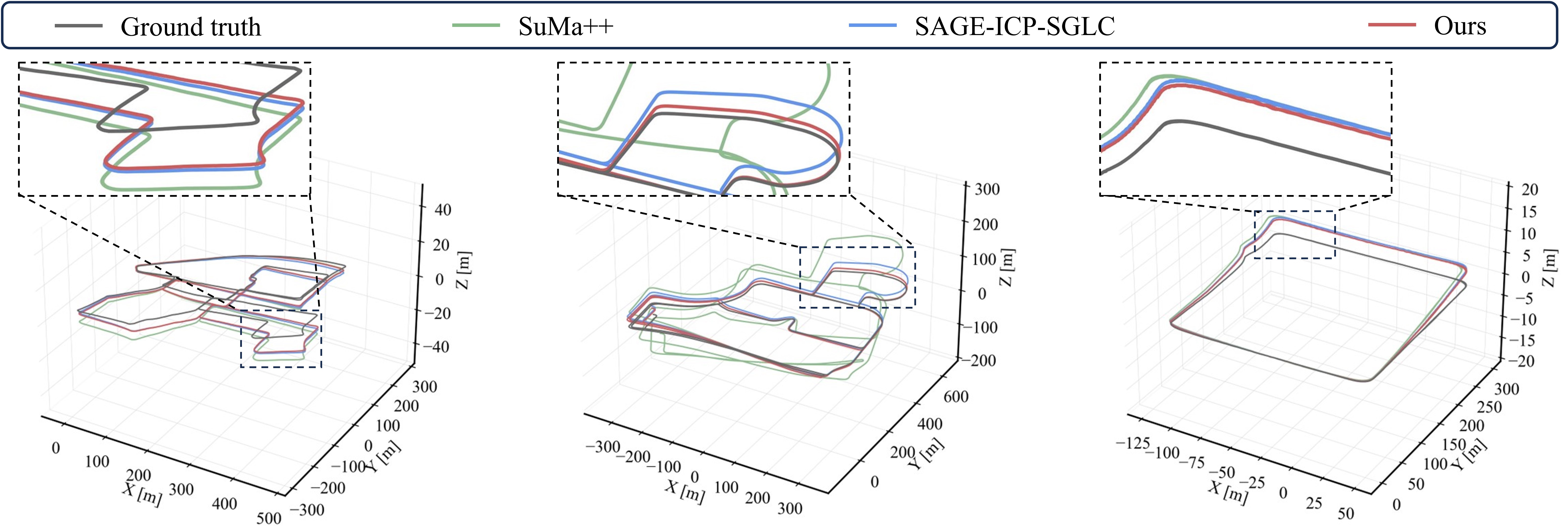}
	\caption{The 3D trajectories analysis on three distinct datasets: KITTI seq.00 (left), MulRAN seq.KA01 (middle), Apollo seq.00 (right), with a comparison to semantic-aided methods.
	}
	\label{fig:traj}
\end{figure*}

To provide a more comprehensive and intuitive demonstration of the precision in our method for pose estimation, we conduct trajectory analysis across three datasets and compare it with SuMa++, SAGE-ICP-SGLC.
As depicted in~\figref{fig:traj}, the SLAM trajectory generated by our method exhibits the closest alignment with the ground truth.

\subsubsection{Odometry performance evaluation}
To exclusively evaluate the odometry performance, we temporarily deactivate the loop closure detection and pose graph optimization components of our method, denoted as Ours-LO, and compare the relative translational error with these semantic-aided methods.
We also additionally incorporate a geometry-based method, KISS-ICP~\cite{vizzo2023ral}, as a baseline.
As shown in ~\tabref{tab:odometry}, for the KITTI dataset, our method exhibits slightly inferior performance compared to SAGE-ICP.
However, SAGE-ICP exhibits significant performance degradation on the MulRAN dataset due to the decline in semantic segmentation performance, a situation similarly observed with SuMa++.
In contrast, our method consistently maintains high performance across multiple datasets.
Therefore, considering both accuracy and robustness, our method holds the best practical value.
Compared with KISS-ICP, the semantic-aided strategy presented in~\secref{sec:odometry} demonstrates practical applicability and performance enhancement.

\subsection{Relocalization Performance Analysis}

\begin{table}[t]
	\caption{Relocalization performance analysis on the KITTI dataset with relative translational error in $\%$.}
	\centering
    \footnotesize
	\begin{threeparttable}
	\begin{tabular}{c|C{0.8cm}|C{0.8cm}|C{0.8cm}|C{0.8cm}}
		\toprule
		Method& Full &2/200 & 5/200 & 10/200\\
		\midrule
        KISS-ICP&  0.52 & 0.78 & 3.31 & 10.52\\
        Ours (w/o relocalization) & \textbf{0.51}  & 0.75 & 3.37& 10.39\\
        Ours &  \textbf{0.51} &  \textbf{0.67} & \textbf{1.45} & \textbf{3.43}\\
		\bottomrule
	\end{tabular}

	\end{threeparttable}
	\label{tab:relocalization}
\end{table}
In this section, we conduct the relocalization performance evaluation on the KITTI dataset to demonstrate the effectiveness and necessity of our proposed semantic graph-based relocalization method.
On the KITTI 00 sequence, we randomly remove $r_n$ consecutive frames of LiDAR data at intervals of $r_m$ frames to simulate situations where processor malfunctions or LiDAR sensor failures lead to the loss of $r_n$ sequential observations for odometry estimation.
Based on the varying degrees of challenge posed by the tasks, we devised four separate experimental configurations, with $r_n$/$r_m$ assigned as follows: 0/200 (full), 2/200, 5/200, 10/200.

The results are presented in~\tabref{tab:relocalization}.
Initially, the random loss of two consecutive LiDAR observations exerts minimal influence on the odometry system.
However, as the number of consecutive lost observations escalates, the performance of the odometry system deteriorates precipitously.
If ten consecutive observations are lost at once, the odometry method without relocalization barely functions properly.
In contrast, our method actively mitigates such errors by attempting relocalization to resume pose tracking when the odometry fails, thereby enhancing the robustness of the system.

\begin{figure}[t]
	\centering
	\includegraphics[width=1\linewidth]{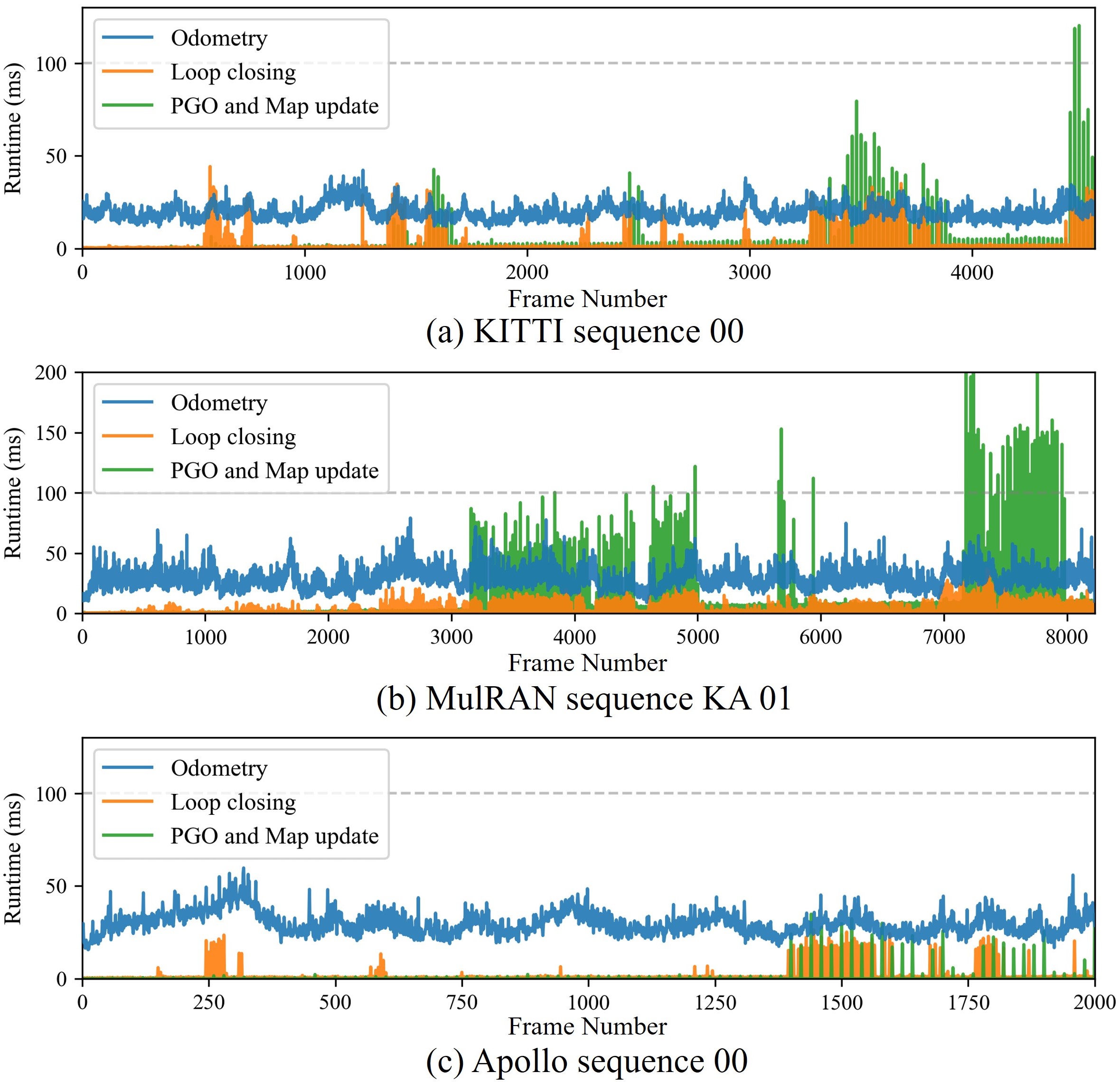}
	\caption{Runtime details on KITTI, MulRAN, and Apollo datasets.
	}
	\label{fig:runtime}
\end{figure}

\subsection{Runtime}
We conduct detailed runtime analysis on three publicly available datasets using a PC equipped with AMD Ryzen 3960X@3.8GHz CPU.
As shown in~\figref{fig:runtime}, in our dual-threaded implementation, the runtime of the odometry thread is relatively stable, averaging approximately 30\,ms.
In the backend thread, loop closing consumes relatively little time, as it is only triggered when a loop closure is detected. A greater proportion of time is allocated to pose graph optimization and map update.
As the number of poses increases, the optimization time gradually increases, yet the average time remains significantly less than 100\,ms.
Additionally, our method spends approximately 12.5\,ms on relocalization.
Overall, our method is fully capable of operating in real-time at a frequency exceeding 10~Hz, thereby highlighting its lightweight characteristics.


\section{Conclusion}
\label{sec:conclusion}
In this paper, we presented an efficient and robust semantic graph-enhanced SLAM framework, which possesses precise pose estimation capabilities while generating both globally consistent semantic graph maps and dense point cloud maps.
The semantic graph, as the cornerstone of our method, facilitates multi-critical functionalities, including relocalization during odometry registration failures, loop closing, and semantic graph map construction.
These designs effectively enhance the accuracy and robustness of SLAM.
Extensive experiments conducted on multiple public datasets demonstrate the superiority of the proposed method.
Additionally, a comprehensive runtime analysis underscores the efficiency of our framework, demonstrating its capability for real-time online operation.

\bibliographystyle{unsrt}
\bibliography{new,glorified}
\end{document}